\documentclass{article}

\usepackage[preprint]{neurips_2024}

\usepackage[utf8]{inputenc} 
\usepackage[T1]{fontenc}    
\usepackage{hyperref}       
\usepackage{url}            
\usepackage{booktabs}       
\usepackage{amsfonts}       
\usepackage{nicefrac}       
\usepackage{xcolor}         

\usepackage{tabularx}
\usepackage{enumitem}
\usepackage{subcaption}
\usepackage{amsmath}
\usepackage{booktabs} 
\usepackage{graphicx}
\usepackage{array} 
\usepackage{subcaption}

\title{EchoAtt: Attend, Copy, then Adjust\\ for More Efficient Large Language Models}

\author{
    Hossein Rajabzadeh\textsuperscript{\rm *,\rm 1,\rm 2},
    Aref Jafari\textsuperscript{\rm *,\rm 1,\rm 2}, 
    Aman Sharma\textsuperscript{\rm 1,\rm 2},
    Benyamin Jami\textsuperscript{\rm 2},\\
    \textbf{Hyock Ju Kwon}\textsuperscript{\rm 1},  
    \textbf{Ali Ghodsi}\textsuperscript{\rm 1},
    \textbf{Boxing Chen}\textsuperscript{\rm 2},
    \textbf{Mehdi Rezagholizadeh}\textsuperscript{\rm 2} \\
    \textsuperscript{\rm 1} University of Waterloo,
    \textsuperscript{\rm 2} Huawei Noah's Ark
    Lab,
    \textsuperscript{\rm *} Equal contributions\\
    \small{\textmd{\{hossein.rajabzadeh, aref.jafari, aman.sharma, hjkwon, ali.ghodsi\}@uwaterloo.ca}} \\
    \small{\textmd{\{mehdi.rezagholizadeh, benyamin.jami, boxing.chen\}@huawei.com}}
}

\begin{document}

\maketitle

\begin{abstract}
Large Language Models (LLMs), with their increasing depth and number of parameters, have demonstrated outstanding performance across a variety of natural language processing tasks. However, this growth in scale leads to increased computational demands, particularly during inference and fine-tuning. To address these challenges, we introduce \textbf{EchoAtt}, a novel framework aimed at optimizing transformer-based models by analyzing and leveraging the similarity of attention patterns across layers. Our analysis reveals that many inner layers in LLMs, especially larger ones, exhibit highly similar attention matrices. By exploiting this similarity, \textbf{EchoAtt} enables the sharing of attention matrices in less critical layers, significantly reducing computational requirements without compromising performance. We incorporate this approach within a knowledge distillation setup, where a pre-trained teacher model guides the training of a smaller student model. The student model selectively shares attention matrices in layers with high similarity while inheriting key parameters from the teacher. Our best results with TinyLLaMA-1.1B demonstrate that \textbf{EchoAtt} improves inference speed by 15\%, training speed by 25\%, and reduces the number of parameters by approximately 4\%, all while improving zero-shot performance. These findings highlight the potential of attention matrix sharing to enhance the efficiency of LLMs, making them more practical for real-time and resource-limited applications.

\end{abstract}

\section{Introduction}
In recent years, Large Language Models (LLMs) have made significant strides in natural language processing (NLP) and extended their reach across a variety of fields\cite{yang2024harnessing, wei2022chain, patil2023gorilla, tahaei-etal-2024-efficient}, revolutionizing applications such as machine translation, text generation, and question answering. The success of these models can largely be attributed to the transformer architecture \cite{vaswani2017attention}, which employs a self-attention mechanism that enables the model to capture contextual relationships between words more effectively than traditional models. However, as the size of these models grows, the computational complexity and memory requirements scale significantly, with a quadratic complexity of $O(n^2)$ for self-attention and $O(n)$ for memory footprint. This growing computational demand creates a bottleneck, particularly during inference and fine-tuning, making these models challenging to deploy in real-time or resource-constrained environments.

Numerous strategies have been proposed to mitigate the computational inefficiency of transformers \cite{zhang2024h2o, rajabzadeh2024qdylora, lieber2024jamba}, including the development of alternative architectures like linear attention models \cite{arora2024simple, yang2023gated,gu2023mamba}. However, these models often struggle to match the generalization and performance capabilities of standard transformer models. Addressing this trade-off between efficiency and performance remains a critical challenge.

In this study, we propose a novel framework to address the inefficiencies of transformer-based LLMs while maintaining their performance. Through an in-depth analysis of attention patterns across different layers of transformers, we observe that in larger models, inner layers tend to exhibit highly similar attention matrices, particularly in generative models. This similarity becomes more pronounced with larger models, aligning with previous findings \cite{ying2021lazyformer,bhojanapalli2021leveraging,he2024matters,liao2024beyond}. Leveraging this insight, we introduce a knowledge distillation-based framework \cite{hinton2015distilling, jafari2021annealing}, which selectively shares attention mechanisms between layers exhibiting high similarity. Our method reduces the number of parameters and computational costs by sharing attention patterns in less critical layers, while retaining unique attention mechanisms in the more distinct layers, typically located in the first and last layers of the network.

To validate our approach, we apply it to the TinyLLaMA-1.1B model and conduct extensive experiments to assess the impact on performance and efficiency. Our results show that by sharing inner attention matrices, we can reduce the parameter count by 3.86\%, while improving inference speed by 15\% and training speed by 25\%. Moreover, this compression comes with minimal loss in accuracy, maintaining competitive performance in zero-shot settings across various benchmarks.

This study not only offers a method for reducing the computational complexity of LLMs but also provides insights into how selective sharing of attention patterns can optimize both the performance and resource efficiency of these models. The contributions of this paper can be
summarized as: 1) introducing \textbf{EchoAtt}, a novel framework designed to optimize transformer-based Large Language Models (LLMs) by leveraging the similarity of attention patterns across layers, 2) proposing a method for \textbf{attention matrix sharing} in less critical layers, significantly reducing computational requirements while maintaining model performance, 3) integrating this approach within a \textbf{knowledge distillation} setup, and 4) demonstrating that \textbf{EchoAtt} improves inference and training speed and also reduces the number of parameters, while maintaining competitive zero-shot performance.

\begin{figure}[t]
\centering

\subfloat{\includegraphics[width = 3.0in]{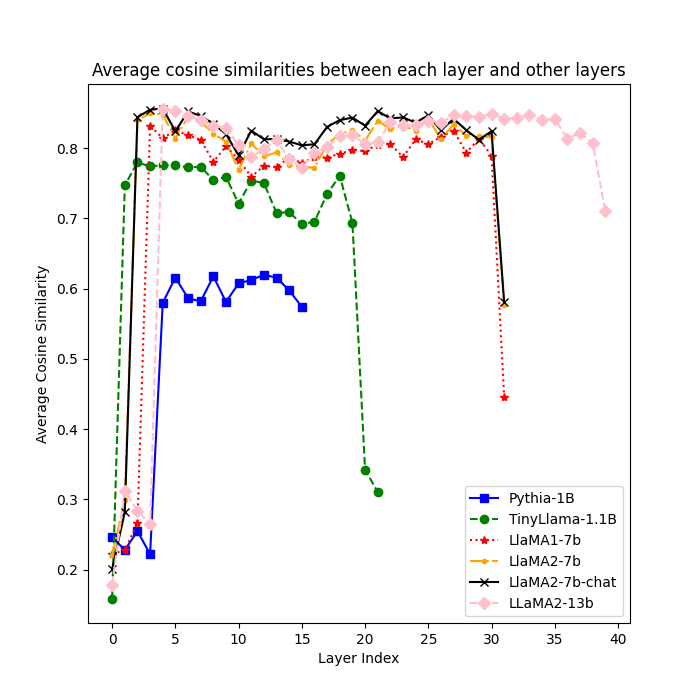}} 
\caption{Average cosine similarities between one layer's attention and other layers' attentions. The results demonstrate that attention scores in some layers are more similar than that of the other layers.}
\label{avg_sim}
\end{figure}

\begin{figure*}[t]
\centering
\subfloat[Pythia-1B]{\includegraphics[width = 1.8in]{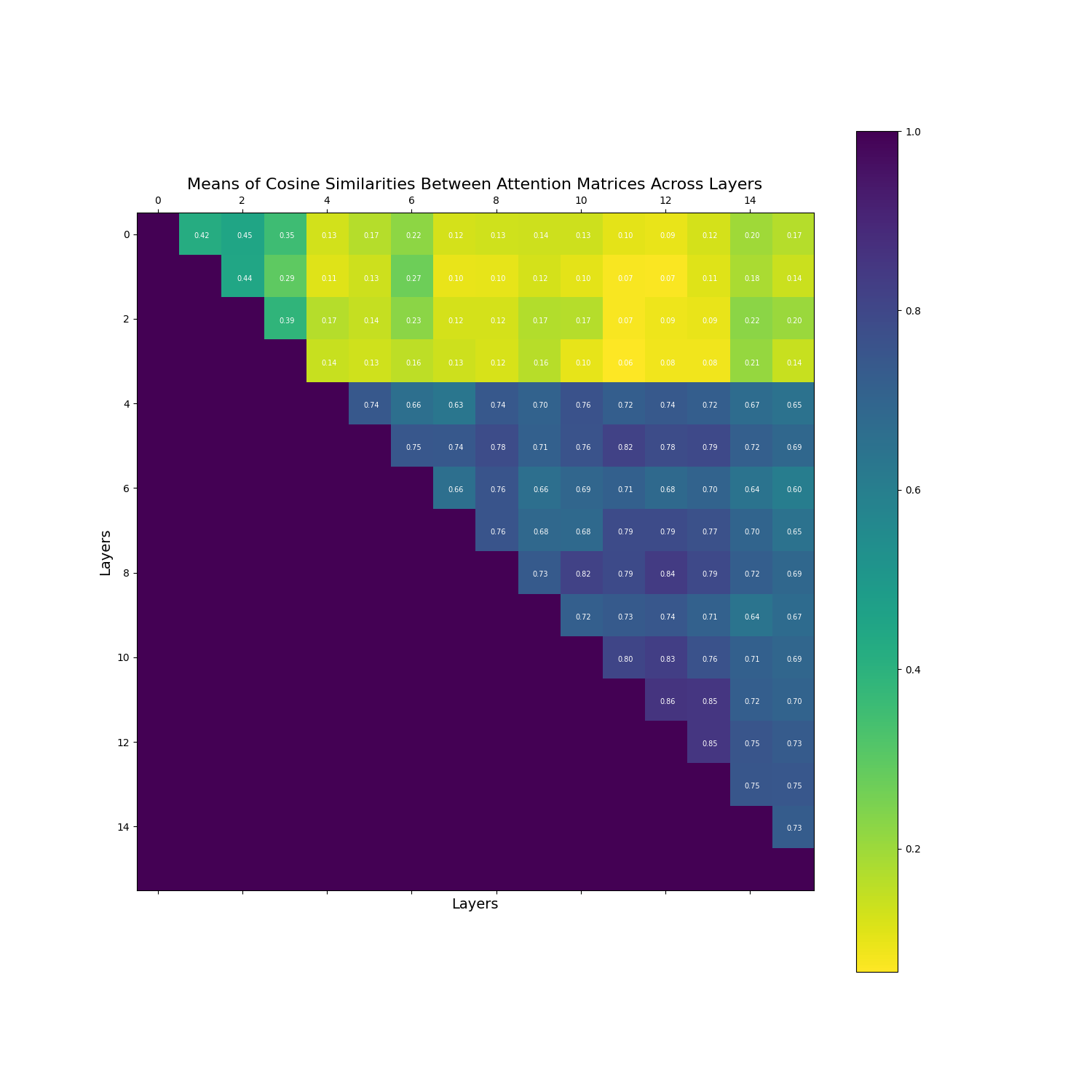}\label{Pythia-1B}} 
\subfloat[TinyLlaMA-1B]{\includegraphics[width = 1.8in]{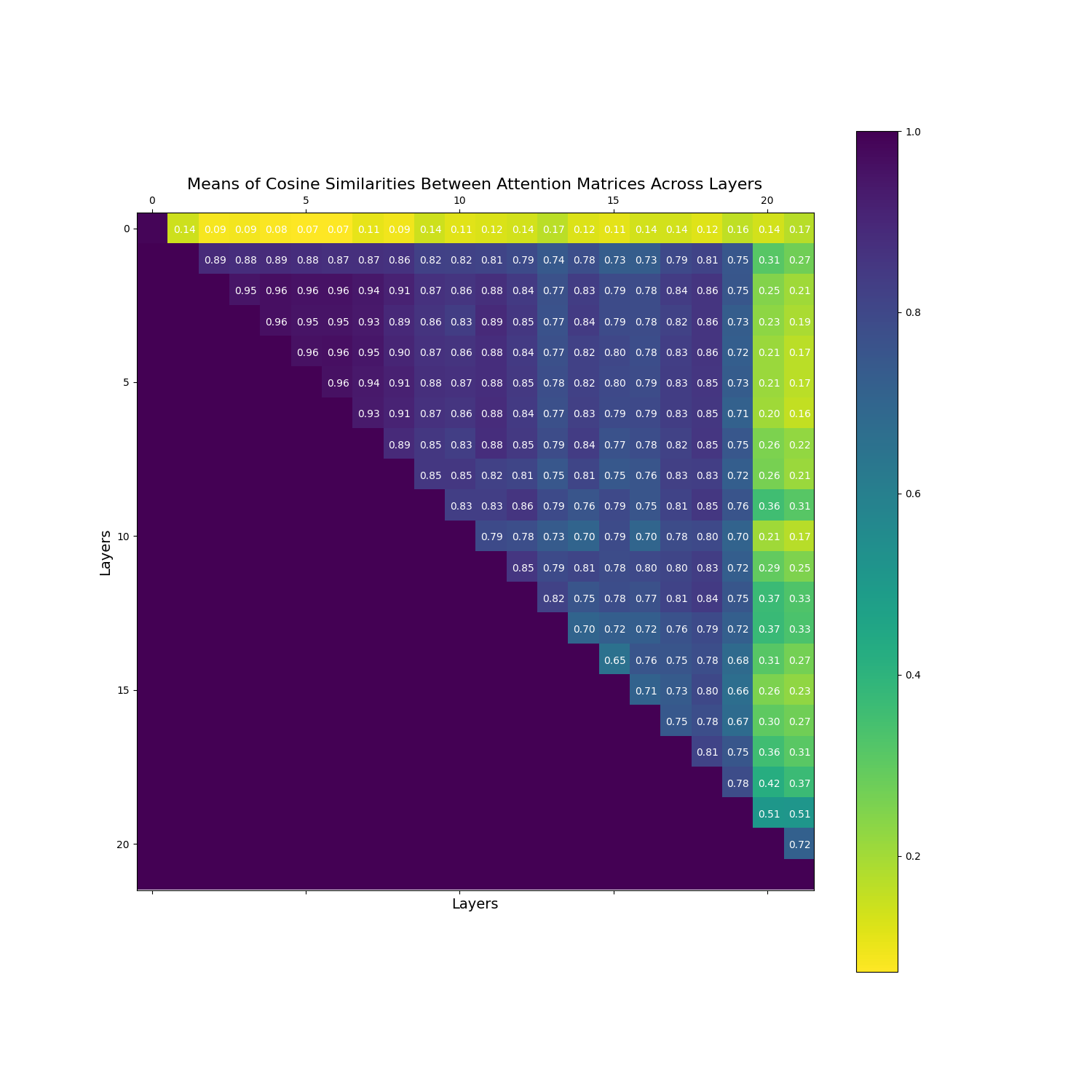}\label{TinyLlaMA-1B}}
\subfloat[LlaMA1-7B]{\includegraphics[width = 1.8in]{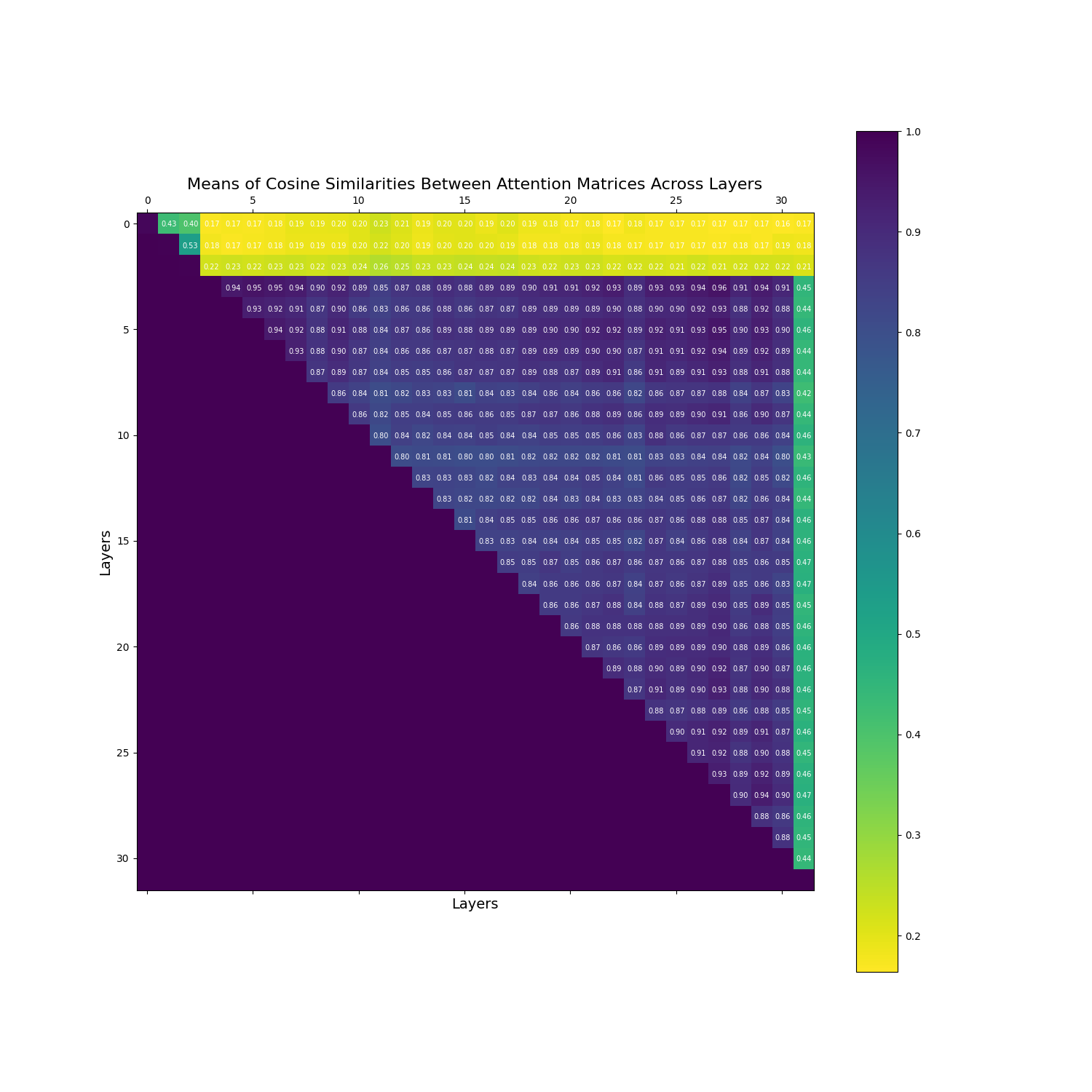}\label{LlaMA1-7B}}\\
\subfloat[LlaMA2-7B]{\includegraphics[width = 1.8in]{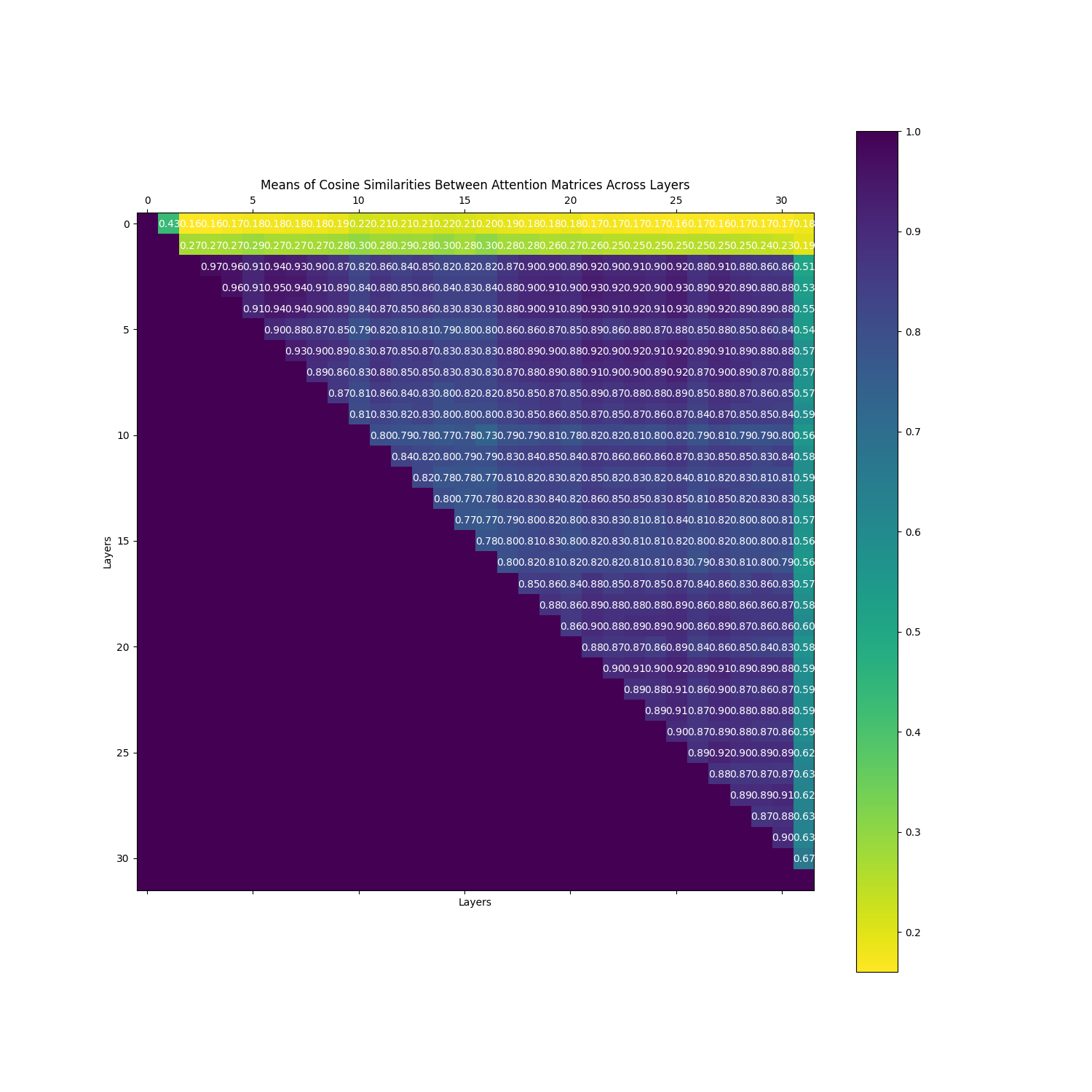}\label{LlaMA2-7BLlaMA2-7B}}
\subfloat[LlaMA2-7B
Chat]{\includegraphics[width = 1.8in]{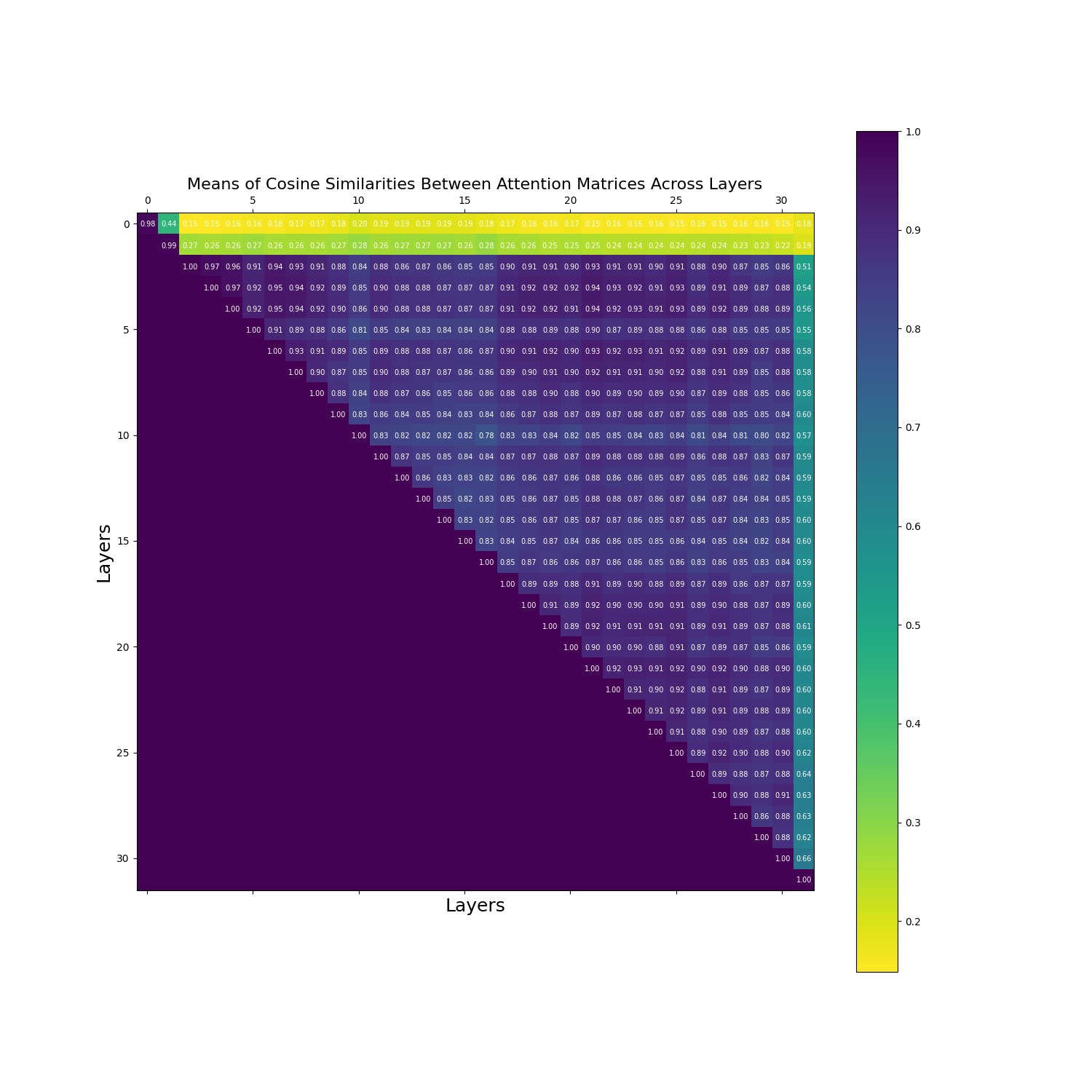}\label{LlaMA2-7B
Chat}}
\subfloat[LlaMA2-13B]{\includegraphics[width = 1.8in]{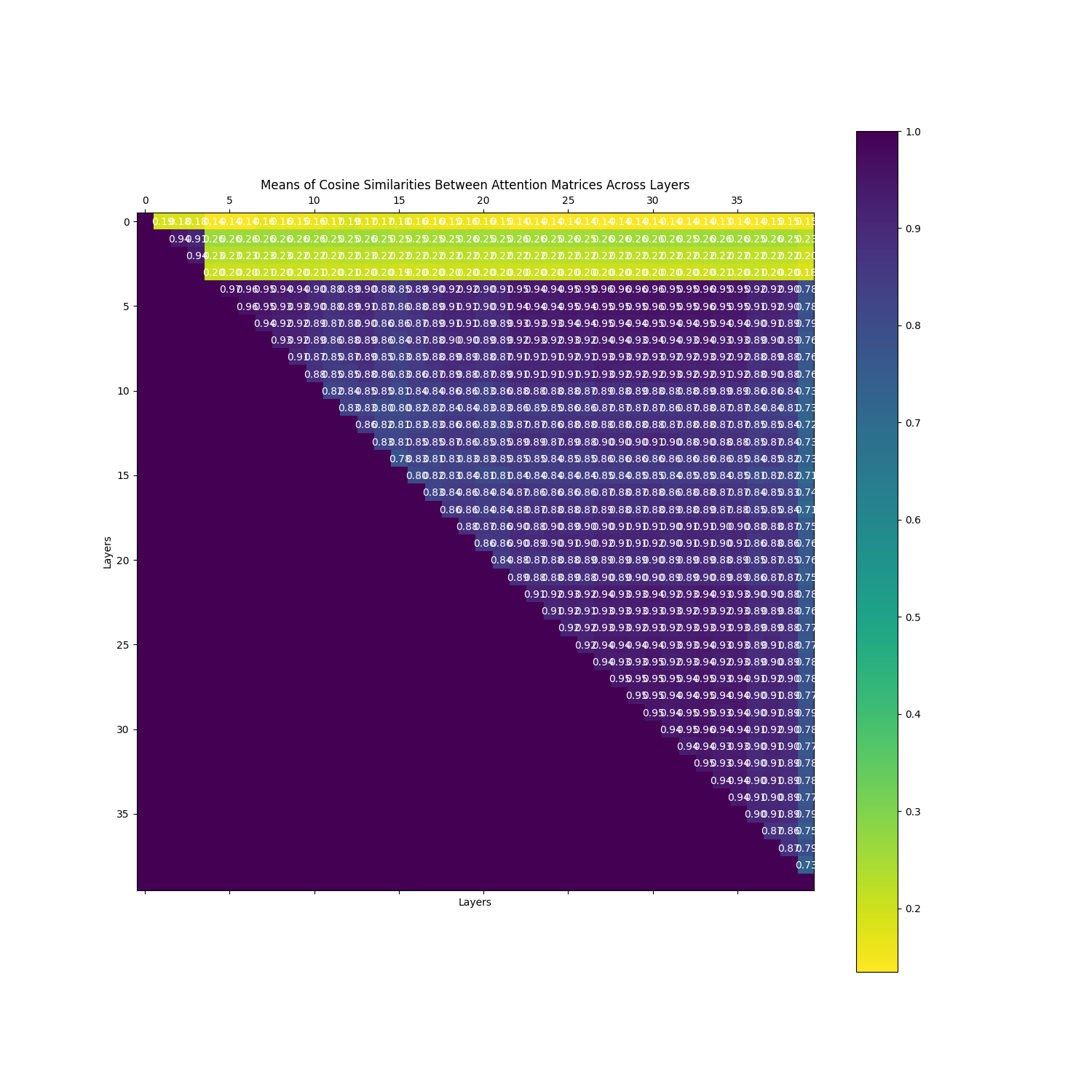}\label{LlaMA2-13B}} 

\caption{Average cosine similarities between the attention matrices of different layers in various LLMs, visualized as upper-triangle matrices. Each entry $[i, j]$ represents the similarity between the attention scores of layer \textit{i} and layer \textit{j}, with higher values indicating more similar attention mechanisms. The results highlight attention similarities in inner layers, suggesting potential for sharing attention mechanisms to reduce computational complexity.}
\label{analysis}
\end{figure*}

\section{Analysis}
\label{sec:analysis}

To analyze the similarity between attention scores across different layers, we employ a subset of the IMDB dataset \cite{maas-EtAl:2011:ACL-HLT2011}. In this subset, each sample is standardized to a length of 512 tokens, effectively eliminating the need for padding and normalizing the sequence length across batches. For each sample, we perform a forward pass and record the attention scores at each layer. We then use cosine similarity as a metric to measure the similarities between the flattened attention scores across the head and embedding dimensions. The results are obtained for several LLMs and illustrated in Figure \ref{analysis}, where each sub-figure depicts the average cosine similarities between attention scores of every pair of layers. For instance, the entry $[i,j]$ in Sub-figure  \ref{TinyLlaMA-1B} and \ref{LlaMA2-13B}, respectively, depict the average cosine similarities between attention scores of layer $i$ and layer $j$ in TinyLlaMA-1B \cite{zhang2024tinyllama} and LlaMA2-13B \cite{touvron2023llama} \footnote{Results are reported in an upper-triangle matrix, as the cosine similarity between attention scores is symmetric, i.e. $cosine(Attn_{l_i},Attn_{l_j})=cosine(Attn_{l_j},Attn_{l_i})$}. Moreover, Figure \ref{avg_sim} demonstrates the average of attention scores between each layer and all other layers. The results indicate that a significant number of layers share similar attention scores, while a few layers, typically the first and last few layers, exhibit distinct attention patterns. This analysis offers a method for identifying which layers produce the most unique attention scores and which layers can potentially share attention mechanisms. For example, in LLaMA2-13B, the first four layers and the last layer demonstrate the most distinctive attention scores, whereas the other layers display more similar attention patterns. 

By identifying layers with highly similar attention scores, we can explore strategies to reduce model complexity, such as attention mechanism sharing. Conversely, recognizing layers with unique attention patterns can guide targeted improvements in model design, ensuring that these critical components are preserved and enhanced. However, naively sharing similar attention scores across layers can lead to performance degradation in shared attention, especially as the number of shared layers increases. To address this issue, the following sections introduce a knowledge distillation mechanism combined with continual training, which significantly enhances the performance of shared attention.

\begin{figure*}[t]
    \centering
    \includegraphics[width=1\textwidth]{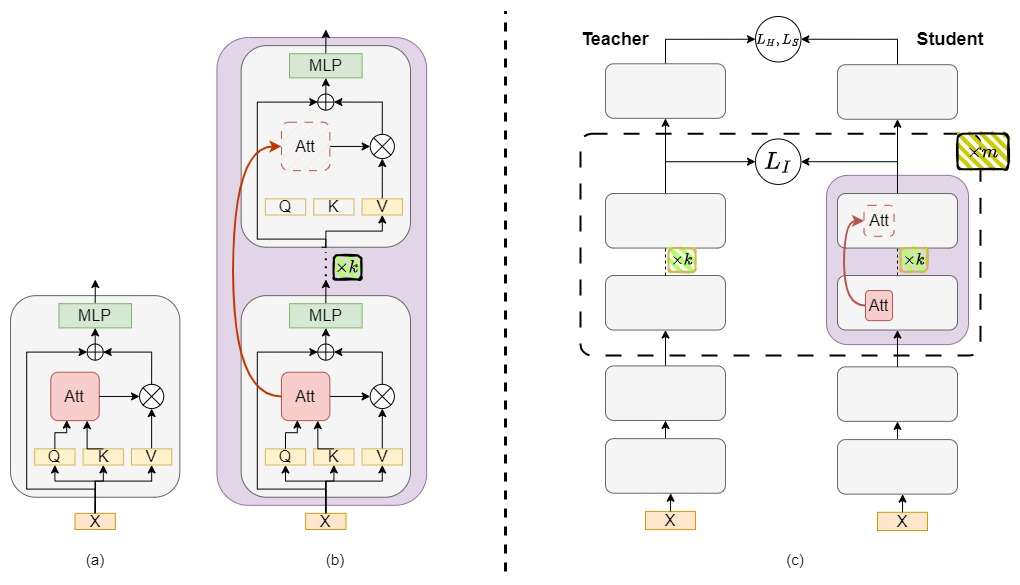}
    \caption{(a) A standard transformer block, which consists of a single transformer layer. (b) A shared attention block, where multiple transformer layers utilize a single attention mechanism. (c) The architecture of the student and teacher models used in the proposed distillation method.}
    \label{fig:diagram}
\end{figure*}

\section{Method: Shared Attention}
The goal of the proposed method is to reduce the computational complexity and memory footprint of transformer-based LLMs by identifying and sharing similar attention patterns across layers. We build upon the observation that many inner layers of these models exhibit highly similar attention matrices, and we propose a framework that leverages this similarity to share attention mechanisms across layers, reducing the number of parameters and computational cost. The method is divided into two main stages: constructing a shared attention student model and applying knowledge distillation from a pre-trained teacher model.

\subsection{Model Construction}

The primary observation driving this work is the high similarity between the attention matrices in the inner layers of transformer models. Specifically, we compute cosine similarity between the attention matrices of different layers and find that many inner layers produce nearly identical attention patterns. Based on this analysis, we design a \textbf{shared attention model} similar to the approach described by \cite{ying2021lazyformer} that optimizes computational efficiency by sharing attention matrices across layers with high similarity.

To construct the student model, we retain the first and last few layers unchanged. To determine which layers to retain, we compute the average cosine similarity of each layer with all other layers, as depicted in Figure \ref{avg_sim}. The layers are then sorted based on their similarity scores, and the cutoff point is set to the maximum distance between similarity scores. Layers with smaller similarity scores are considered unchanged. Additionally, we impose a constraint that the unchanged layers must be among the first or last layers. For the inner layers, we implement a shared attention mechanism similar to the approach described by \cite{ying2021lazyformer}. Specifically, every \( k \) consecutive inner layers are grouped into what we refer to as a \textit{shared attention block} (see Figure \ref{fig:diagram}-b), where attention matrices are shared among these blocks. This design significantly reduces computational time and memory footprint of the student model by avoiding the computation of separate attention matrices for the shared layers and eliminating the \( K \) and \( Q \) values in these layers, with minimal impact on performance. The hyperparameter \( k \) controls the extent of model compression achieved through this design.

\subsubsection{Shared Attention Blocks}

In standard transformers, each layer computes its own attention matrix using the query, key, and value matrices \(Q_i\), \(K_i\), and \(V_i\). In the proposed shared attention model (see Figure \ref{fig:diagram}b), for each shared attention block, a single set of \(Q\) and \(K\) matrices is computed and used for all layers within the block. This reduces the number of parameters and avoids redundant computations of highly similar attention matrices.

Formally, the attention mechanism in standard transformers is defined as:

\[
\text{Att}_i = \text{softmax}\left(\frac{Q_i K_i^T}{\sqrt{d}}\right)V_i
\]

\noindent where $i$ is the layer index.

In our shared attention model, for every \(k\) consecutive layers with similar attention matrices, we compute a shared attention mechanism as:

\[
\text{A}_{shared} = \text{softmax}\left(\frac{Q_{shared} K_{shared}^T}{\sqrt{d}}\right)
\]

\[
\text{Att}_j = A_{shared}V_j, \hspace{0.5cm} j \in [i, i+k]
\]

\noindent where $i$ is the shared attention block index.

\subsubsection{Parameter Sharing Strategy}

The parameter-sharing mechanism is controlled by a hyperparameter \(k\), which dictates the number of consecutive layers that share attention matrices. A larger value of \(k\) results in more aggressive parameter sharing and greater model compression, while smaller values of \(k\) preserve more unique attention patterns. 

\subsection{Knowledge Distillation}

Once the shared attention student model is constructed, we employ a \textbf{knowledge distillation} approach to transfer knowledge from a pre-trained teacher model to the student model. This process helps recover performance lost due to the parameter sharing and ensures that the student model achieves competitive results.

\subsubsection{Distillation Setup}

The knowledge distillation process consists of two stages:

\begin{itemize}
    \item \textbf{Stage 1: Distillation with teacher's Pseudo-Labels}  
   In the first stage, the student model is trained using the outputs of the teacher model as pseudo-labels. Both the student and teacher models are fed the same input tokens, and the student is trained to match the teacher’s output at multiple levels. Three loss functions are used to guide the distillation process (see Figure \ref{fig:diagram}c):
   \begin{itemize}
       \item \textbf{Intermediate Layer Loss (\(\mathcal{L}_I\))}: This loss aligns the intermediate layer outputs of the student and teacher models for each shared attention block. We use mean squared error to minimize the distance between the shared layers of the student and the corresponding layers of the teacher.
       \[
       \mathcal{L}_I = \frac{1}{m} \sum_{i=1}^{m} \| S_{ki+b}(x) - T_{ki+b}(x) \|_2^2
       \]
      where \( m \) represents the number of shared attention blocks, \( k \) denotes the number of attention layers within each shared block, and \( b \) indicates the number of early layers that are skipped. \( S_j \) and \( T_j \) refer to the outputs of the student and teacher models at layer \( j \).
      
       \item \textbf{Soft Label Loss (\(\mathcal{L}_S\))}: A KL-divergence loss is used to match the soft label distributions of the student and teacher models. This loss ensures that the student learns from the probability distributions produced by the teacher model.
       \[
       \mathcal{L}_S = \text{KL}(\sigma(S(x)) \| \sigma(T(x)))
       \]
       \item \textbf{Hard Label Loss (\(\mathcal{L}_H\))}: Cross-entropy loss is applied to distill hard labels sampled from the teacher model into the student model. This step helps the student model learn from the teacher's confident predictions.
       \[
       \mathcal{L}_H = \text{CE}(\sigma(S(x)), \tau(T(x)))
       \]
   \end{itemize}

   The functions \( \sigma \) and \( \tau \) denote the softmax and argmax functions, respectively. and $S(x)$ and $T(x)$ are student and teacher models outputs.
   
   The final loss function is a weighted combination of these three components:
   \[
   \mathcal{L} = \alpha \mathcal{L}_I + \beta \mathcal{L}_S + \gamma \mathcal{L}_H
   \]
   where \( \alpha \), \( \beta \), and \( \gamma \) are tunable coefficients controlling the contribution of each loss function.
   
   \item \textbf{Stage 2: Refinement with True Labels}  
   In the second stage, the student model is further fine-tuned using the actual labels from the training dataset. This stage allows the student to refine its predictions and improve its accuracy. Cross-entropy loss is used for this step, and the student is trained to directly predict the true labels from the input data.
\end{itemize}

\section{Evaluations and Results}
\subsection{Experimental Setup} 
For all training experiments, we employed a subset of the Slim-Pajama dataset \cite{cerebras2023slimpajama}, comprising over 3.7 billion tokens. During the knowledge distillation and continual training stages, the models were trained for 1 epoch and 0.25 epochs, respectively, on this dataset. A detailed list of the critical hyper-parameters used in our experiments is provided in Table \ref{params} in Appendix \ref{hyperparameters}. It is important to note that no hyper-parameter fine-tuning was applied during the experiments. The hyper-parameters were kept consistent across all stages of training to ensure that the results reflect the true performance of the models under identical conditions, without any optimization specific to individual tasks or datasets. Additionally, we used LLaMA-Factory for training \cite{zheng2024llamafactory} and LM-Evaluation-Harness \cite{eval-harness} for evaluation.

\begin{table*}[t]
  \centering
  \caption{Performance comparison of the Shared-Attention TinyLlaMA model across different attention-sharing
ratios in a zero-shot evaluation. The table presents accuracy metrics obtained under continual training conditions \textbf{without} any knowledge distillation. Baseline results are compared against three variations of the model with 77\%,
41\%, and 23\% attention-sharing ratios.}
  \label{tab:tinyllama_comparison1}
  \small 
  \scalebox{0.99}{
  \begin{tabularx}{\textwidth}{>{\hsize=0.5\hsize}X>{\hsize=0.625\hsize}X>{\hsize=0.725\hsize}X>{\hsize=0.525\hsize}X>{\hsize=0.525\hsize}X}
    \toprule
    Benchmark & TinyLlaMA(baseline) & Ours (77\%) & Ours (41\%) & Ours (23\%) \\
    \midrule
    mmlu & $25.27\pm0.37$ & $23.44\pm0.36$ & $25.05\pm0.37$ & $25.43\pm0.50$ \\
    winogrande & $59.35\pm1.38$ & $50.75\pm0.14$ & $54.54\pm0.14$ & $58.41\pm1.39$ \\
    swag & $51.69\pm0.35$ & $46.29\pm0.35$ & $49.80\pm0.35$ & $51.0\pm0.35$ \\
    hellaswag & $46.40\pm0.05$ & $36.57\pm0.49$ & $43.35\pm0.49$ & $44.50\pm0.50$ \\
    xnli\_en & $41.53\pm0.99$ & $45.22\pm0.01$ & $52.64\pm0.01$ & $52.61\pm0.01$ \\
    agieval\_en & $17.03\pm0.01$ & $16.77\pm0.01$ & $17.39\pm0.01$ & $17.19\pm0.01$ \\
    \hline
    \textbf{Average} & 40.21 & 36.50 & 39.51 & \textbf{41.52} \\
    \bottomrule
  \end{tabularx}
  }
\end{table*}

\subsection{Results} 
To validate the efficacy of our model, we employ TinyLlaMA \cite{zhang2024tinyllama}\footnote{\url{https://huggingface.co/TinyLlama/TinyLlama_v1.1}} as our baseline LLM. Tables \ref{tab:tinyllama_comparison1} and \ref{tab:tinyllama_comparison2}  compares the accuracy of the baseline against its shared attention versions where a certain percentage of attention layers are shared across the network, indicated by the sharing ratios 77\%, 41\%, and 23\%. Table \ref{tab:tinyllama_comparison1}   demonstrates the shared attention performance with just continual training while Table \ref{tab:tinyllama_comparison2} repeats the same experiments with both knowledge distillation and continual training. The results show that with continual training only, the model with a 23\% sharing ratio outperforms the baseline, the 41\% ratio performs comparably to the baseline, and the 77\% ratio underperforms. However, when shared attention is combined with both knowledge distillation and continual training, the models with 23\% and 41\% sharing ratios outperform the baseline, while the performance gap for the model with 77\% sharing is significantly reduced. The superior performance of the models with 23\% and 41\% sharing ratios could be attributed to a slight regularization effect of shared attention, as noted by \cite{bondarenko2024low}. In this context, sharing attention leads to a reduction in the number of parameters, which may act as a form of regularization, thereby enhancing model performance.

Overall, the results indicate that knowledge distillation coupled with continual training improves the performance of shared attention in all sharing ratios.

\begin{table*}[t]
  \centering
  \caption{Performance comparison of the Shared-Attention TinyLlaMA model across different attention-sharing ratios in a zero-shot evaluation. The table presents accuracy metrics obtained under continual training conditions, \textbf{coupled with} knowledge distillation. Baseline results are compared against three variations of the model with 77\%, 41\%, and 23\% attention-sharing ratios, demonstrating the impact of varying the degree of attention sharing on overall model performance.}
  \label{tab:tinyllama_comparison2}
  \small 
  \scalebox{0.99}{
  \begin{tabularx}{\textwidth}{>{\hsize=.5\hsize}X>{\hsize=0.625\hsize}X>{\hsize=0.725\hsize}X>{\hsize=0.525\hsize}X>{\hsize=0.525\hsize}X}
    \toprule
    Benchmark & TinyLlaMA (baseline) & Ours (77\%) & Ours (41\%) & Ours(23\%) \\
    \midrule
    mmlu  &$25.27\pm0.37$ &$24.55\pm0.36$ & $25.96\pm0.37$ & $25.94\pm0.50$ \\
    winogrande & $59.35\pm1.38$ & $52.33\pm1.40$ & $58.01\pm1.39$ & $58.36\pm1.4$ \\
    swag & $51.69\pm0.35$ & $47.96\pm0.35$ & $50.33\pm0.35$ & $50.78\pm0.35$ \\
    hellaswag & $46.40\pm0.05$ & $38.56\pm0.49$ & $43.82\pm0.49$ & $44.40\pm0.50$ \\
    xnli\_en & $41.53\pm0.99$ & $49.32\pm1.0$ & $53.25\pm1.0$ & $52.97\pm0.01$ \\
    agieval\_en &$17.03\pm0.01$ &$17.50\pm0.59$ & $17.94\pm0.59$ & $17.21\pm0.01$ \\
    \hline
    \textbf{Average} & 40.21 & 38.37 & \textbf{41.55} & \textbf{41.61} \\
    \bottomrule
  \end{tabularx}
  }
\end{table*}

\begin{table*}[t]

  \centering
  \caption{Comparing the baseline TinyLlaMA-1.1B against its shared attention versions in terms of speedup in training, inference, and the reduced portion of parameters. The inference speed is reported based on one 32GiG-V100 GPU, and the training speed is computed by eight 46GiG-L40-GPUs.}
  \label{tab:tinyllama_comparison}
  \small 
  \begin{tabularx}{\textwidth}{>{\hsize=0.99\hsize}X>{\hsize=0.725\hsize}X>{\hsize=0.725\hsize}X>
  {\hsize=0.725\hsize}X}
    \toprule
    Model &Inference Speed &Training Speed  &Reduced Parameters  \\
    \midrule
     TinyLlaMA (baseline) &28.44 token/sec &43h,30m  &None \\
     \hline
    Shared TinyLlaMA(23\%) &30.99 (9\%faster)  &37h,30m (14\%faster)  &24 millions (2.14\%) \\
    Shared TinyLlaMA(41\%) &32.61 (15\%faster) &32h,50m (25\% faster)  &43 millions (3.86\%) \\
    Shared TinyLlaMA(77\%) &40.50 (42\%faster)  &23h,30m (46\%faster) &80 millions (7.29\%)  \\
    \hline
    \bottomrule
  \end{tabularx}
  \label{speedup-gain}
\end{table*}

Table \ref{speedup-gain}
compares the baseline TinyLLaMA-1.1B model with its shared attention versions, focusing on inference speed, training speed, and the reduction in the number of parameters. The shared attention models demonstrate notable improvements in both inference and training speeds. Specifically, the model with 77\% shared attention achieves the highest performance gains, with a 42\% increase in inference speed, reaching 40.50 tokens per second, compared to the baseline's 28.44 tokens per second. In terms of training efficiency, this same model reduces the training time by 46\%, completing the process in 23 hours and 30 minutes, down from the baseline's 43 hours and 30 minutes. Additionally, the shared attention models also exhibit a reduction in the total number of parameters. The 77\% shared attention model reduces the number of parameters by $\approx$ 80 million, corresponding to a 7.29\% reduction. The other shared models follow a similar trend, with the 23\% and 41\% shared attention models achieving reductions of 24 million (2.14\%) and 43 million (3.86\%) parameters, respectively. Overall, these results indicate that the proposed approach of shared attention not only accelerates both inference and training processes but also leads to a more parameter-efficient model, making it a promising technique for optimizing large language models.

\subsection{Ablation Study}
This ablation study aims to determine whether continual training can enhance the performance of our baseline model. To investigate this, we subjected our baseline model, TinyLlaMA without attention sharing, to continual training using the same dataset and settings as the shared attention models. The results, presented in Table \ref{ablation_1}, indicate that continual training not only fails to improve TinyLlaMA's performance but actually leads to a slight decrease in its average performance. Therefore, continual training does not benefit the vanilla TinyLlaMA model and may even be detrimental under the conditions tested.

\begin{table*}[t]

  \centering
  \caption{Zeros-shot evaluation of  TinyLlaMA before and after continual training. No attention sharing is applied here. The results demonstrate a slight decrease in average performance, suggesting that continual training may not be beneficial for this model under the tested conditions.}
  \small 
  \begin{tabularx}{\textwidth}{>{\hsize=0.8\hsize}X>{\hsize=0.725\hsize}X>{\hsize=0.825\hsize}X}
    \toprule
    Benchmark & TinyLlaMA-1.1B (baseline) & Continual trained TinyLlaMA  \\
    \midrule
    mmlu &$25.27\pm0.37$ & $25.39\pm0.37$ \\
    winogrande & $59.35\pm1.38$ & $58.41\pm1.38$ \\
    swag & $51.69\pm0.35$ & $51.44\pm0.32$ \\
    hellaswag & $46.40\pm0.05$ & $46.10\pm0.05$ \\
    agieval\_en &$17.03\pm0.01$ & $16.98\pm0.59$ \\
    TruthfulQA\_mc1 &$22.28\pm1.50$ & $20.69\pm1.40$ \\
    TruthfulQA\_mc2 &$35.09\pm1.40$ & $34.08\pm1.40$ \\
    \hline
    \textbf{Average} & 36.69 & 36.15 \\
    \bottomrule
  \end{tabularx}
  \label{ablation_1}
\end{table*}

\begin{table*}[t]
  \centering
  \caption{Zero-shot evaluation of the Distilled-Shared LlaMA-160m in terms of accuracy before and after continual training. Shared attention ratio is set to 33\%. The results indicate that continual training, when combined with knowledge distillation, outperforms the approach where continual training is omitted from the shared attention training process.}
  \label{ablation_21}
  \small 
  \begin{tabularx}{\textwidth}{>{\hsize=0.8\hsize}X>{\hsize=0.725\hsize}X>{\hsize=0.825\hsize}X>
  {\hsize=0.825\hsize}X}
    \toprule
    Benchmark & LlaMA-160m (baseline) & Distilled-Shared-Attn &Continual-Distilled-Shared-Attn  \\
    \midrule
    mmlu &23.02 &22.97  &22.97 \\
    winogrande &50.12  &51.70  &51.54 \\
    swag &40.21 &36.39  &38.91 \\
    hellaswag &30.94  &29.32 &29.92  \\
    agieval\_en &17.52 &16.95 &17.86 \\
    \hline
    \textbf{Average} & 32.36 & 31.46 & 32.24 \\
    \bottomrule
  \end{tabularx}
\end{table*}

The next ablation study evaluates the impact of continual training on top of distillation stage based on the performance of shared attention models. To that end, we employ LlaMA-160m \cite{miao2023specinfer} and, first, train the shared attention version of this model with a sharing ratio of $33\%$ (the indices of shared layers are: [4,6,8,10]) while excluding the continual training stage. Let's call this model as Distilled-Shared-Attn. Then, we allow the Distilled-Shared-Attn to receive the continual training stage, and refer to this model by Continual-Distilled-Shared-Attn. Table \ref{ablation_21} demonstrates the performance of these two models and compare them with the baseline, i.e. LlaMA-160m. The results clearly show that the shared attention models achieve competitive performance compared to the baseline. Notably, the performance of the shared attention models further improves when continual training is applied on top of knowledge distillation, demonstrating the effectiveness of this approach in enhancing model accuracy and generalization. This observation highlights the potential of shared attention mechanisms in reducing model complexity without compromising performance, especially when combined with advanced training techniques.

\section{Conclusion}
We investigated attention mechanisms in large language models and proposed a framework that identifies and shares less important attentions, coupled with knowledge distillation and continual training to recover performance. Our experiments demonstrated that, on TinyLLaMA-1.1B, this approach improved average zero-shot performance, increased training and inference speeds up to 42\% and 46\%, respectively, and reduced parameters by 7.29\%. These results highlight the effectiveness of selective attention sharing in enhancing model efficiency without compromising performance.

\medskip

{
\small

\bibliographystyle{unsrtnat}
\bibliography{references}





\appendix


 

\section{Hyper-Parameters}
\label{hyperparameters}
The complete list of hyper-parameters used in our experiments is detailed in Table \ref{params}. It is important to note that no hyper-parameter fine-tuning was applied during the experiments. The hyper-parameters were kept consistent across all stages of training to ensure that the results reflect the true performance of the models under identical conditions, without any optimization specific to individual tasks or datasets.

\begin{table*}[htbp]

  \centering
  \caption{List of essential hyperparameters used in the experiments, detailing the key settings that governed model training.}
  \small 
  \begin{tabularx}{\textwidth}{>{\hsize=0.79\hsize}X>
  {\hsize=0.995\hsize}X}
    \toprule
    Name &Value   \\
    \midrule
     Optimizer &AdamW  \\
     Deepspeed stage &Zero3 \\
     Learning rate (lr) &1e-4  \\
     Lr scheduling type &Cosine \\
     Max sequence length &2048  \\
     Global batch size &1024  \\
     FP16 &True  \\
     Warmup ratio &0.005  \\
     $[\alpha,\beta,\gamma]$ & $[0.25,0.25,0.5]$ \\
    LlaMA-160m: Shared layer indices (33\%) &[4,6,8,10] out of 12 layers \\
     TinyLlaMA: Shared layer indices (23\%) &[2,5,4,3,7] out of 22 layers\\
     TinyLlaMA: Shared layer indices (41\%) &[2,5,4,3,7,6,18,9] out of 22 layers\\
     TinyLlaMA: Shared layer indices (77\%) &[2,5,4,3,7,6,18,9,8,11,12,1,17,10,14,13,16] out of 22 layers\\
     
    \hline
    \bottomrule
  \end{tabularx}
  \label{params}
\end{table*}

\section{Limitation}
Despite the promising results, our study has certain limitations that need to be considered. First, due to computational constraints, we were unable to evaluate the performance of shared attention mechanisms on larger language models (LLMs) \footnote{However, our analysis suggests that larger LLMs tend to exhibit similar attention patterns across layers, which could indicate that shared attention mechanisms might be particularly effective in these models as well.}. Extending our experiments to encompass models with greater parameter counts would provide deeper insights into the scalability and effectiveness of our approach in more complex architectures. Such an evaluation could reveal potential challenges or benefits that are not apparent in smaller models like TinyLLaMA-1.1B.

Second, we did not investigate how supervised fine-tuning (SFT) operates within the shared attention framework for downstream tasks. Exploring the interaction between SFT and shared attention models could offer valuable information on how these models perform when adapted to specific applications, such as question answering. 

\end{document}